\documentclass[10pt,twocolumn,letterpaper]{article}

\usepackage{cvpr}              % To produce the CAMERA-READY version
\usepackage[dvipsnames]{xcolor}

\usepackage{times}
\usepackage{epsfig}
\usepackage{graphicx}
\usepackage{amsmath}
\usepackage{amssymb}

\usepackage{subcaption}
\usepackage{enumitem}
\usepackage{multicol}
\usepackage{multirow}
\usepackage{booktabs}
\usepackage{algorithm}
\usepackage{algorithmic}
\usepackage{pgfplots}
\usepackage{colortbl}

\usepackage[accsupp]{axessibility}

\definecolor{cvprblue}{rgb}{0.21,0.49,0.74}
\usepackage[pagebackref,breaklinks,colorlinks,citecolor=cvprblue]{hyperref}

\title{Multi-View Attentive Contextualization for Multi-View 3D Object Detection}

\author{
Xianpeng Liu$^{1}$, Ce Zheng$^{2}$, Ming Qian$^{3}$, Nan Xue$^{3}$, Chen Chen$^{2}$, Zhebin Zhang$^{4}$, Chen Li$^{4}$, Tianfu Wu$^{1}$ \\
$^1$North Carolina State University \quad $^2$University of Central Florida \\
$^3$Ant Group \quad $^4$OPPO U.S. Research Center\\
\url{https://xianpeng919.github.io/mvacon}
}

\begin{document}
\maketitle
\begin{abstract}

We present Multi-View Attentive Contextualization (MvACon), a simple yet effective method for improving 2D-to-3D feature lifting in query-based multi-view 3D (MV3D) object detection. Despite remarkable progress witnessed in the field of query-based MV3D object detection, prior art often suffers from either the lack of exploiting high-resolution 2D features in dense attention-based lifting, due to high computational costs, or from insufficiently dense grounding of 3D queries to multi-scale 2D features in sparse attention-based lifting. 
Our proposed MvACon hits the two birds with one stone using a representationally dense yet computationally sparse attentive feature contextualization scheme that is agnostic to specific 2D-to-3D feature lifting approaches.  
In experiments, the proposed MvACon is thoroughly tested on the nuScenes benchmark, using both the BEVFormer and its recent 3D deformable attention (DFA3D) variant, as well as the PETR, showing consistent detection performance improvement, especially in enhancing performance in location, orientation, and velocity prediction. 
It is also tested on the Waymo-mini benchmark using BEVFormer with similar improvement. 
We qualitatively and quantitatively show that global cluster-based contexts effectively encode dense scene-level contexts for MV3D object detection.
The promising results of our proposed MvACon reinforces the adage in computer vision \--- ``(contextualized) feature matters". 

\end{abstract}    

\setlength{\belowdisplayskip}{1pt} \setlength{\belowdisplayshortskip}{1pt}
\setlength{\abovedisplayskip}{1pt} \setlength{\abovedisplayshortskip}{1pt}

\section{Introduction}
\label{sec:intro}

\begin{figure}
     \centering
     \begin{subfigure}[b]{0.235\textwidth}
         \centering
         \includegraphics[width=\textwidth]{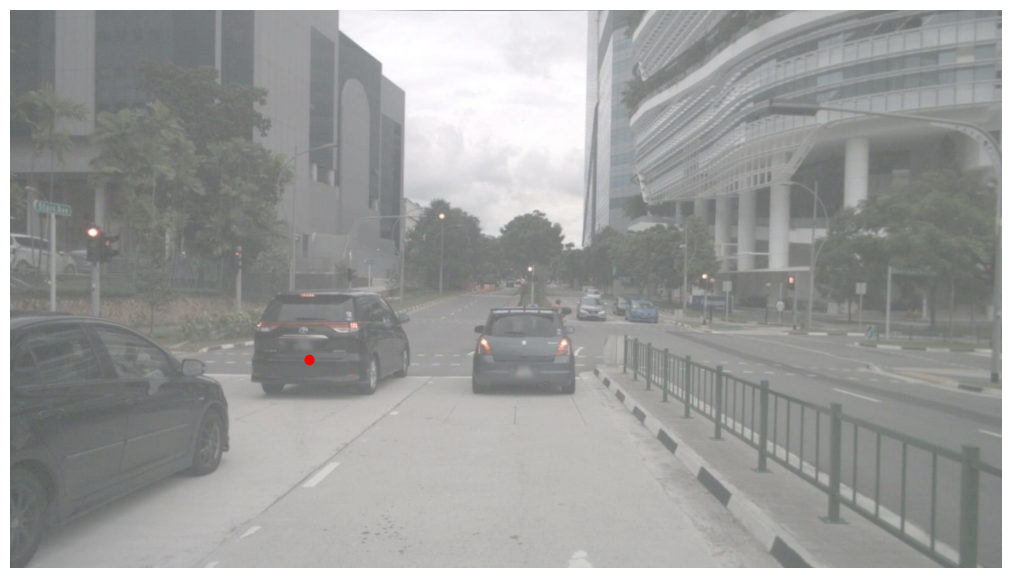}
         \caption{Projected BEV anchor}
         \label{fig:bev_anchor}
     \end{subfigure}
     \hfill
     \begin{subfigure}[b]{0.235\textwidth}
         \centering
         \includegraphics[width=\textwidth]{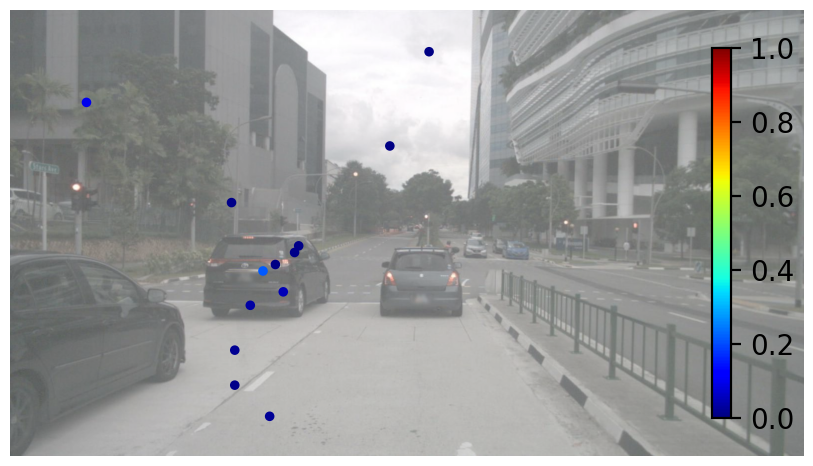}
         \caption{BEVFormer \cite{bevformer}}
         \label{fig:bevformer_pts}
     \end{subfigure}
     \hfill
     \\
     \begin{subfigure}[b]{0.235\textwidth}
         \centering
         \includegraphics[width=\textwidth]{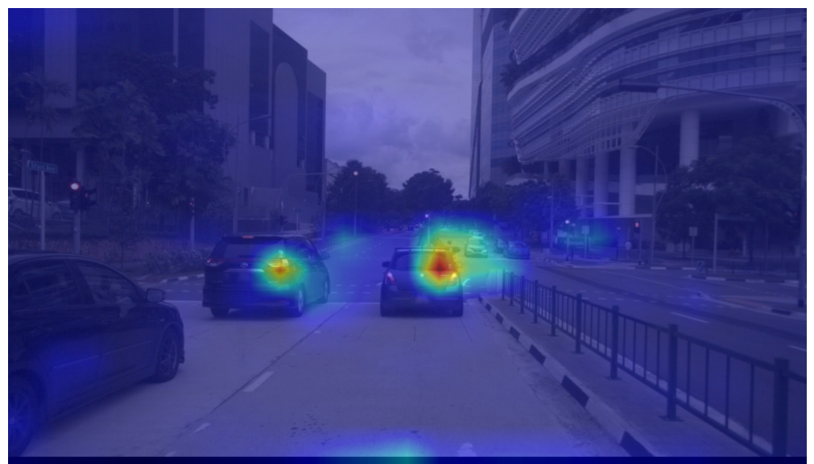}
         \caption{Learned clusters}
         \label{fig:heatmap}
     \end{subfigure}
     \hfill
     \begin{subfigure}[b]{0.235\textwidth}
         \centering
         \includegraphics[width=\textwidth]{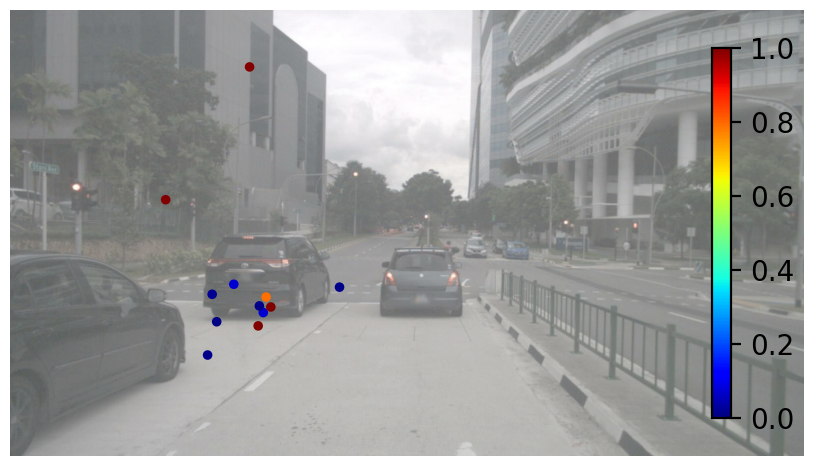}
         \caption{Our MvACon}
         \label{fig:paca_pts}
     \end{subfigure}
    \vspace{-2em}
    \caption{\small The effects of our proposed MvACon in the 2D-to-3D feature lifting. Consider a point (red) on the car in (a), which is projected from a 3D BEV anchor point. In lifting 2D features to ground the 3D BEV anchor point, vanilla BEVFormer~\cite{bevformer} utilizes a predefined number of deformable points, with offsets learned through a 6-layer cross-attention module relative to the projection point. (b) shows the deformed points after the final cross-attention layer, most of which have low attention weights, indicating the model's uncertainty or inability to consolidate contributions effectively for good lifting. Our MvACon tackles this issue with clustering-based attention, as visualized in (c). (d) shows the deformed points, where we observe not only high-confidence points on the car but also on the building. We further observe that the points on the building remain stable across encoding layers (see Fig.~\ref{fig:dfpts}) and consecutive frames (see suppl.).  With those high-confident deformed points in a spatiotemporally stable configuration, our MvACon may induce a local object-context aware coordinate system that helps the overall performance, especially the estimation of velocity and orientation, as we quantitatively observed in experiments. See text for details.}
    \label{fig:teaser}\vspace{-4mm}
\end{figure}

Camera-based 3D object detection is a pivotal area of research in computer vision, particularly owing to its appealing applications in cost-effective autonomous systems, such as autonomous driving and robot autonomy. Recently, significant progress has been witnessed in the field of multi-view 3D (MV3D) object detection, especially with the advent of end-to-end MV3D detection approaches~\cite{bevsurvey}. \textit{One crucial component in the end-to-end MV3D detection system is the 2D-to-3D feature lifting module}, which converts perspective 2D multi-view image feature maps to 3D feature representation. It aims to counter the complete loss of depth information in individual 2D images by exploiting multi-view clues. However, a significant challenge in practical applications like autonomous driving is the often insufficient field-of-view overlap across views, making it difficult to effectively address the loss of depth information.

To perform 2D-to-3D feature lifting, recent methods \cite{huang2021bevdet, bevformer, wang2022detr3d, petr} aim to learn a unified 3D space representation using 3D anchors that are either sparsely or uniformly sampled. These methods generally fall into two categories subject to the interaction of 3D anchors with 2D features and the feature aggregation strategy.
(1) The Lift-Splat-Shoot (LSS) method~\cite{lss, huang2021bevdet, huang2022bevdet4d, li2023bevdepth, li2023bevstereo} first lifts 2D features into 3D (pseudo-LiDAR) space using the outer product with the estimated depth, then assigns them to the nearest 3D anchors.
(2) In contrast, the query-based design~\cite{wang2022detr3d, bevformer, petr}, pioneered by the DETR method~\cite{detr} for end-to-end 2D object detection, adopts 3D anchors as queries and uses 2D image features as keys and values. They interact and aggregate via spatial cross-attention in the expressive Transformer architecture~\cite{vaswani2017attention}. 
These two paradigms have been widely used in downstream tasks like map segmentation \cite{liu2023petrv2} and occupancy prediction \cite{bevformer}. \textit{This paper primarily focuses on the query-based detection paradigm.} One reason is that LSS-based methods often encounter excessive computational complexity and issues with error propagation and depth estimation magnification post-lifting, potentially capping their performance. However, the query-based design also grapples with heavy computation costs or limited 3D information awareness, depending on their Transformer design.
 
In this paper, we focus on addressing limitations of two main paradigms of query based MV3D object detectors in a unified way (elaborated in Sec. \ref{approach}). In particular, we introduce multi-view attentive contextualization (MvACon) to address  limitations of decoder-only dense attention methods like PETR~\cite{petr}, which lack high-resolution features due to computational constraints, and to simultaneously address the issue of sparsely grounded 3D anchors in encoder-decoder 2D/3D deformable attention methods such as BEVFormer~\cite{bevformer} and DFA3D~\cite{dfa3d}. \textit{Our proposed MvACon aims to be representationally dense while computationally sparse}. To achieve this, we expand the conventional three-component paradigm of MV3D object detection to a four-component setup (Fig.~\ref{fig:pipeline}): (1) 2D image representation learning through a feature backbone shared across views, (2) \textit{MvACon for attentive contextualization of the 2D features}, (3) 2D-to-3D feature lifting, and (4) a 3D object detection head or decoder that utilizes these lifted features. This modular design allows our MvACon to remain agnostic to specific 2D-to-3D feature lifting strategies and aligns with the classic adage in representation learning and computer vision: ---‘(contextualized) feature matters’. 

More specifically, our approach contextualizes original feature maps extracted from the backbone network using a cluster-attention operation. This builds upon the recently proposed patch-to-cluster attention (PaCa)~\cite{grainger2023paca}. For perspective-based decoder-only detectors like PETR, we apply cluster contextualization before the feature maps are fed into the decoder. For encoder-decoder based detectors, such as BEVFormer and DFA3D, we incorporate cluster contextualization within the spatial cross-attention operation. Through extensive experiments, we demonstrate that our proposed MvACon effectively and consistently enhances query-based MV3D object detectors by encoding more useful contexts, thereby facilitating better 2D-to-3D feature lifting. Rigorously controlled experiments reveal that, for perspective-based decoder-only detectors, the cluster attention contextualization significantly improves localization and velocity prediction. In the case of encoder-decoder based detectors, it effectively reduces errors in location, orientation, and velocity.

In summary, our main contributions are:
\begin{itemize}
    \item We analyze and address the limitation of 2D-to-3D feature lifting in the prior art, that is the lack of sufficient 3D representational power due to their local 3D awareness. 
    \item We propose MvACon (Multi-view Attentive Contextualization) to induce the global 3D awareness in an easy-to-integrate way to enhance the 2D-to-3D feature lifting in both decoder-only based MV3D object detectors and encoder-decoder based ones. 
    \item We show consistent performance improvement of our MvACon on the challenging NuScenes~\cite{caesar2020nuscenes} dataset using three baseline query-based MV3D object detectors, as well as on the Waymo-mini~\cite{waymo} benchmark.
\end{itemize}

\section{Related Work}
\label{sec:related_work}

\textbf{Camera-based 3D Object Detection.}
Camera-based 3D object detection can be primarily categorized into two settings: single-view and multi-view. In the realm of monocular 3D object detection research, addressing the challenge of inaccurate object localization~\cite{monodle} is critical. Researchers have exerted considerable effort to utilize monocular depth cues. This includes transforming inputs into pseudo-lidar point clouds~\cite{pseudolidar, patchnet, am3d} and explicitly incorporating depth into models~\cite{pseudostereo, d4lcn, monodtr}. Another significant research direction involves the explicit use of geometric priors, encompassing approaches like key-point constraints~\cite{li2020rtm3d, chen2022epro, liu2021autoshape}, shape projection relationships~\cite{gupnet, deep3dbox, dcd, monoflex}, and temporal depth estimation~\cite{dfm}. Innovations in monocular 3D object detection also include novel loss modules~\cite{monopair, monodis, dimembed}, 3D-aware backbones~\cite{deviant, kumar2021groomed, m3drpn}, and second-stage detection paradigms~\cite{did, monoxiver}. In multi-view settings, the configuration often closely resembles that of monocular setups due to the limited field-of-view overlap between the different camera views. Therefore, MV3D object detection focuses on addressing the challenge of learning universal representation for multi-view sensors. It has benefited from advancements in various techniques such as view lifting~\cite{lss, wang2022detr3d, bevformer, petr, vedet, dfa3d}, depth encoding~\cite{li2023bevdepth, li2023bevstereo, 3dppe, dfa3d}, and temporal modeling~\cite{bevformer, huang2022bevdet4d, liu2023petrv2, park2022time, lin2022sparse4d, streampetr}. In our multi-view approach, we aim at addressing the challenge of multi-view representation learning with focus on enhancing the view lifting module within query-based detection methods by empowering original features with cluster-based contextual features.

\vspace{0.2em}\noindent\textbf{Representation Learning in Camera-based 3D Object Detection.}
Camera-based 3D object detection is inherently a data-intensive task due to its ill-posed nature, the expansive search space in 3D, and the scarcity of labeled data in scenes. Consequently, developing robust representations for this task is both critical and challenging. Early research in monocular 3D object detection has demonstrated the utility of depth contexts~\cite{dd3d, dd3dv2} and projection contexts~\cite{monocon} in enhancing detection capabilities. Recent advances also highlight the effectiveness of scene-level representations, such as density fields~\cite{mildenhall2021nerf}, in improving 3D representation learning~\cite{xu2023mononerd}. In the domain of multi-view research, most leading methods utilize backbones pre-trained with projection contexts (e.g., FCOS3D~\cite{wang2021fcos3d} weights) or depth contexts (e.g., DD3D~\cite{dd3d} weights). However, these pre-trained weights may not fully leverage the capabilities of newer backbone network designs. Recent studies have begun to explore alternatives to this pre-trained paradigm, including the integration of an auxiliary projection context branch in end-to-end training~\cite{bevformerv2}. Our work aims to enhance network representation by explicitly incorporating scene-level cluster context as supplementary information during the view lifting stage in query-based MV3D object detectors.

\vspace{0.2em}\noindent\textbf{Vision Transformers.}
Since the pioneer work of ViT~\cite{vit}, extensive research~\cite{vitsurvey} has been dedicated to enhancing the representational abilities of neural networks for visual tasks. It has been established that CNNs and Transformers can mutually augment each other’s capabilities, as evidenced in designs like Transformer-enhanced CNNs~\cite{vt1, vt2} and CNN-enhanced Transformers~\cite{vt3, vt4, vt5}. Additionally, a significant branch of visual Transformer research focuses on developing new attention mechanisms tailored to the locality bias in vision tasks. Notable examples include HaloNet~\cite{vt6}, SWin~\cite{liu2021swin}, Deformable Attention~\cite{deformabledetr}, and VOLO~\cite{vt7}, all introducing innovative local attention mechanisms to mitigate the quadratic computational cost associated with visual inputs. Concurrently, models like TNT~\cite{vt8}, ViL~\cite{vt9}, PVTv2-linear~\cite{wang2022pvt}, POTTER~\cite{zheng2023potter}, and PaCa~\cite{grainger2023paca} explore the integration of local and global contexts. Inspired by the progress of Vision Transformers, our work addresses the limitations of two prevalent paradigms in query-based MV3D detectors caused by their attention mechanisms.
\section{Approach}
\label{approach}

\begin{figure*}[ht]
    \centering
    \includegraphics[width=0.98\textwidth]{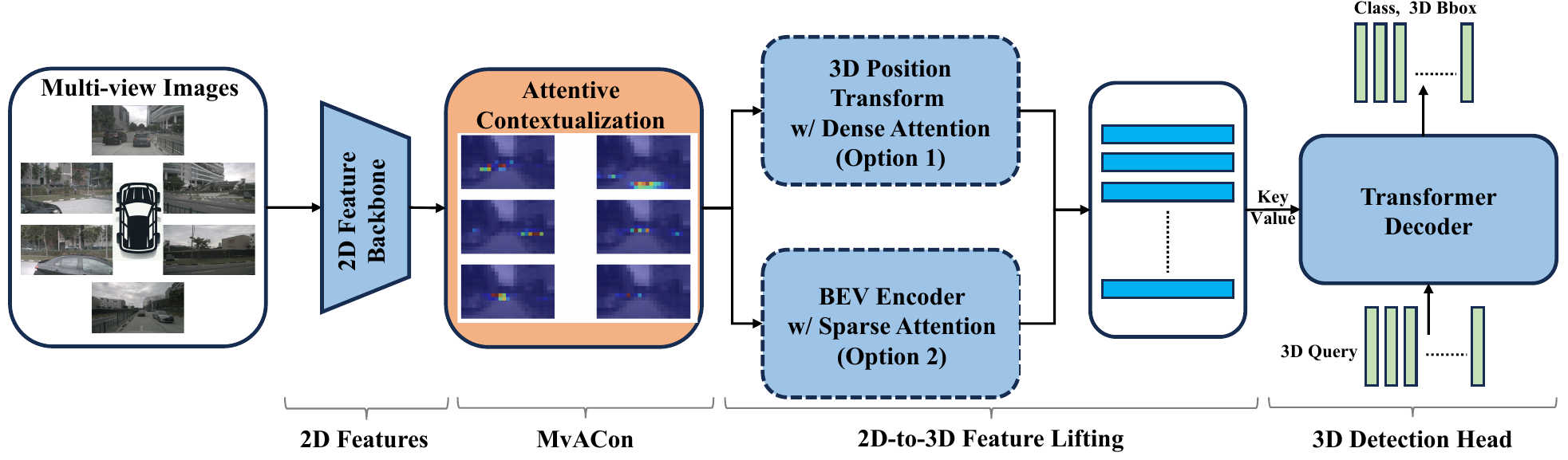}
    \vspace{-1em}
    \caption{{Overview of a query-based MV3D object detection pipeline with our proposed MvACon.} Our proposed MvACon is a plug-and-play module for two state-of-the-art query-based MV3D object detection paradigms (e.g., PETR~\cite{petr} and BEVFormer~\cite{bevformer} respectively), which computes attentively contextualized features to facilitate better 2D-to-3D feature lifting in the two paradigms. See text for details. 
    }
    \label{fig:pipeline}\vspace{-2mm}
\end{figure*}

Given a set of images {$I_i \in \mathbb{R}^{3 \times H \times W}$} from $N$ cameras with known extrinsics $T_i \in \text{SE}(3)$ and intrinsics $K_i \in \mathbb{R}^{3 \times 3}$, MV3D object detection aims to infer the label (e.g., Car, Pedestrian, Barrier) and the 3D bounding box for each object instance in the scene. 
In this section, we first delve into the pipeline of query-based MV3D object detection in Sec.~\ref{sec:overview}. We then analyze the pros and cons of the core 2D-to-3D feature lifting component in two state-of-the-art MV3D object detection methods in Sec.~\ref{sec:pro}. Finally, we present our proposed MvACon in Sec.~\ref{sec:mvacon}.

\subsection{Query-based MV3D Object Detection}
\label{sec:overview}

For better understanding, we explain the query-based MV3D object detection pipeline, shown in Fig.~\ref{fig:pipeline}, in a reverse manner. The 3D detection head typically builds upon DETR3D \cite{wang2022detr3d}, which is based on the original DETR \cite{detr}. Initially, it defines a sufficiently large number, $O$, of latent $C$-dimensional 3D object queries, $Q_{O,C}$. These object queries update using Keys and Values derived from multi-view inputs, followed by a classification head predicting object labels and a bounding box regression head determining the 3D bounding boxes. Obviously, the key challenge lies in how to transform / lift multi-view 2D inputs into 3D-aware Keys and Values.

To this end, a feature backbone is trained to extract deep 2D features from the multi-view input images. The complexity arises from different design choices for feature lifting. It mainly involves two aspects: representation and computation. From the representational perspective, multi-scale feature pyramids, crucial in 2D object detection, become even more essential in 3D object detection. Computationally, handling multi-view inputs is already demanding. Adding multi-scale feature pyramids without careful optimization can significantly increase the computational load. There are two strategies in the state-of-the-art development of query-based MV3D object detection.

\textbf{Decoder-Only Architectures: Single-Scale Multi-View 2D Features with Dense Attention.} These designs are among the most straightforward. They involves using multi-view 2D feature maps from the last layer of the feature backbone, which are then concatenated and flattened along the spatial dimensions to form Keys and Values. The latent object queries, $Q_{O,C}$, are updated using a vanilla Transformer (i.e., each object query attends to every 2D location in the multi-view inputs, termed dense attention). However, this basic approach often fails as it does not encode any 3D-aware information. To address this, The PETR \cite{petr} introduces a physically-meaningful 3D position transformation as positional encoding, added to the multi-view 2D feature maps before concatenation and flattening. It has proven effective for query-based MV3D object detection, as illustrated in Option 1 in Fig.~\ref{fig:pipeline}).

\textbf{Encoder-Decoder Architectures: Multi-Scale Multi-View 2D Features to Latent BEV Queries with Sparse Attention.} The BEV (Bird's Eye View) representation acts as a unified, grid-based and ego-centric scene representation with predefined grid sizes (e.g., $200\times 200$) on the XZ plane. Geometrically, the BEV grid can be treated as a pillar-based point-cloud representation, collapsed along the Y-axis. A predefined number of points along the pillar (Y-axis) direction in the BEV grid will be uniformly sampled. These points form a uniform geometry prior for the underlying 3D scene and serve as BEV anchors to elevate 2D features into the BEV space. With known camera poses, these sampled points can be projected to each view at multiple scales. However, due to the uniform geometry prior in the projection, the projected points require deformability to better align with data observations. The BEVFormer \cite{bevformer} addresses this by introducing a sparse deformable attention mechanism (see Option 2 in Fig.~\ref{fig:pipeline}). To counter the uniform geometry prior further, it learns a small predefined number of offsets, rather than directly deforming the projected points on each view, to lift 2D features from those deformed points with attentive weights. Latent BEV queries are introduced in learning these offsets and attentive weights. The BEV encoder's role is to refine the BEV queries, enabling them to provide meaningful offsets and attentive weights for lifting 2D features to 3D BEV anchors. Finally, embedded BEV queries as Keys/Values in the 3D detection head (i.e., decoder) update the latent object queries, $Q_{O,C}$, e.g., through sparse deformable attention as in the BEVFormer, before predicting the 3D object detection results.

\subsection{The Limitation of 2D-to-3D Feature Lifting in the Prior Art}
\label{sec:pro}

Although they both have shown remarkable progress for MV3D object detection, the PETR pipeline and the BEVFormer pipeline have a common limitation in their 2D-to-3D feature lifting, that is the local or shallow 3D awareness, rather than the desirable counterpart, the global and semantic meaningful 3D awareness.  

In the PETR pipeline, consider the 2D feature map of the $n$-th view, $F^n_{h\times w\times c}$, where $(h,w)$ are the spatial sizes, height and width respectively, and $c$ the feature dimension of the backbone. The 3D position transform converts the (shared) camera frustum discretized as a mesh grid of sizes $(h, w, D)$ to the 3D space based on the known camera poses, where $D$ is the discretized depth levels. After the conversion, each 3D point is represented by a normalized 3D coordinate in the homogeneous form, i.e., $(x,y,z,1)$. So the 3D position transformation results in the positional encoding $P^n_{h\times w\times 4\cdot D}$. Both $F^n_{h\times w\times c}$ and $P^n_{h\times w\times 4\cdot D}$ are projected into the space of the same dimensionality, $d$ using a linear layer and a Multi-layer Perceptron (MLP) with the ReLU nonlinearity respectively, we have $F^n_{h\times w\times d}$ and  $P^n_{h\times w\times d}$ which are summed in an element-wise way. Consider the latent representation of the 3D position in $P^n_{h\times w\times d}$ with the depth grid fused by the MLP, it is grounded to one feature point in $F^n_{h\times w\times d}$, leading to the local 3D awareness. 

In the BEVFormer pipeline, as we show in Fig.~\ref{fig:teaser}, the projected BEV anchor is often  grounded to some low-confidence and scattered deformed points. Although the grounding may not be spatially local, they are often semantically shallow. 

\subsection{Our Proposed MvACon Method}
\label{sec:mvacon}

Our goal is to address the local or shallow 3D awareness stated above by introducing an easy-to-integrate (mostly plug-and-play)  module to learn the global and semantically meaningful 3D-awareness, as illustrated in Fig.~\ref{fig:pipeline}.
The basic idea of our MvACon is to attentively contextualize the 2D features in the 2D-to-3D lifting. 

In the PETR pipeline, our idea is to contextualize the individual 2D feature map, $F^n_{h\times w\times d}$, 
\begin{equation}
    \mathbf{F}^n_{h\times w\times d} = \text{MvACon}(F^n_{h\times w\times d}), 
\end{equation}
where after the contextualization every feature point in $\mathbf{F}^n_{h\times w\times d}$ can connect to the entire map $F^n_{h\times w\times d}$, inducing the global 3D awareness for grounding the positional encoding $P^n_{h\times w\times d}$.

In the BEVFormer pipeline, our idea is to contextualize the multi-scale feature maps, e.g., the $l$-th layer of the feature pyramid of the $n$-th view, $F^{n,l}_{h\times w\times c}$, 
\begin{equation}
    \mathbf{F}^{n,l}_{h\times w\times c} = \text{MvACon}(\{F^{n,l}_{h\times w\times c}\}_{l=1}^L), \label{eq:mvacon_bev}
\end{equation}
where after the contextualization every feature point in $\mathbf{F}^{n,l}_{h\times w\times c}$ can connect to the entire $L$-layer feature pyramid $\{F^{n,l}_{h\times w\times c}\}_{l=1}^L$, inducing the global 3D awareness for grounding  projected BEV anchors on the $n$-the view.

To achieve the global contextualization effect, we adapt the recently proposed Patch-to-Cluster attention (PaCa)~\cite{grainger2023paca} method. The core idea of PaCa is to leverage a learnable clustering module to cluster a feature map into a predefined number $M$ of clusters. For notional simplicity, consider a feature map $F_{h\times w\times c}$ as $N=h\times w$ tokens $F_{N\times c}$, the clustering assignment is computed by,
\begin{equation}
    \mathcal{C}_{N,M} = \text{Softmax}(\text{Clustering}(F_{N,c})),
\end{equation}
where $\text{Clustering}()$ can be implemented in different ways (see our ablation studies in Tab.~\ref{Tab:cluster_methods}), and the $\text{Softmax}$ is along the token dimension. Then, we compute $M$ clusters by,
\begin{equation}
    z_{M,c} = \text{LN}(\mathcal{C}^{\top}_{N,M}\cdot F_{N,c}),
\end{equation}
where $\text{LN}()$ is the layer normalization \cite{ln}. 

Then, the PaCa-based MvAcon is defined by,
\begin{equation}
    F'_{N,c} = \text{Softmax}(\frac{Q_{N,c}\cdot K_{M,c}^{\top}}{\sqrt{c}})\cdot V_{M,c} + F_{N,c},
\end{equation}
where $Q_{N,c}$ is the linear projection of $F_{N,c}$, $K_{M,c}$ and $V_{M,c}$ are from the clusters $z_{M,c}$. The second term is the shortcut. The multi-head PaCa can be straightforwardly defined. For Eqn.~\ref{eq:mvacon_bev}, we concatenate the clusters from all the pyramid layers before computing the Key and the Value. Here, the PaCa module is of linear complexity.

\section{Experiments}

\begin{table*}[t]
\begin{center}
\vspace{-0.5em}
\resizebox{0.75\linewidth}{!}{
\begin{tabular}{l|c|ccccc|c}
\toprule
\textbf{Method} & \textbf{mAP}$\uparrow$ & \textbf{mATE}$\downarrow$ & \textbf{mASE}$\downarrow$ & \textbf{mAOE}$\downarrow$ & \textbf{mAVE}$\downarrow$ & \textbf{mAAE}$\downarrow$ & \textbf{NDS}$\uparrow$ \\

\midrule

PETR-VovNet-99~\cite{petr} & 37.8 & 74.6 & 27.2 & \textbf{48.8} & 90.6 & \textbf{21.2} & 42.6 \\

PETR-VovNet-99-MvACon & \textbf{38.2 (+0.5)} & \textbf{73.9} & \textbf{27.0} & 50.5 & \textbf{84.1} & 21.3 & \textbf{43.4 (+0.8)} \\

\midrule

BEVFormer-t~\cite{bevformer} & 25.2 & 90.0 & 29.4 & 65.5 & 65.7 & \textbf{21.6} & 35.4 \\

BEVFormer-t-MvACon & \textbf{25.9 (+0.7)} & \textbf{88.4} & \textbf{28.8} & \textbf{64.6} & \textbf{60.5} & 22.5 & \textbf{36.5 (+1.1)}\\

\midrule
 
BEVFormer-s~\cite{bevformer} & 37.0 & 72.1 & 28.0 & 40.7 & 43.6 & 22.0 & 47.9 \\

BEVFormer-s-MvACon & \textbf{39.3 (+2.3)} & \textbf{71.3} & \textbf{27.7} & \textbf{40.1} & \textbf{42.0} & \textbf{19.7} & \textbf{49.6 (+1.7)} \\

\midrule

BEVFormer-b~\cite{bevformer} & 41.6 & 67.3 & \textbf{27.4} & 37.2 & 39.4 & \textbf{19.8} & 51.7 \\

BEVFormer-b-MvACon & \textbf{42.6 (+1.0)} & \textbf{66.4} & 27.6 & \textbf{35.0} & \textbf{36.2} & 20.0 & \textbf{52.8 (+1.1)} \\

\midrule

DFA3D-s~\cite{dfa3d} & \textbf{40.1} & 72.1 & 27.9 & 41.1 & 39.1 & \textbf{19.6} & 50.1 \\

DFA3D-s-MvACon & \textbf{40.1} & \textbf{71.0} & \textbf{27.4} & \textbf{38.3} & \textbf{37.2} & 20.8 & \textbf{50.6 (+0.5)} \\

\midrule

DFA3D-b~\cite{dfa3d} & {43.0} & \textbf{65.4} & \textbf{27.1} & 37.4 & 34.1 & \textbf{20.5} & 53.1 \\

DFA3D-b-MvACon & \textbf{43.2 (+0.2)} & 66.4 & 27.5 & \textbf{34.4} & \textbf{32.3} & {20.7} & \textbf{53.5 (+0.4)}\\

\bottomrule
\end{tabular}
}
\end{center}
\vspace{-1.5em}
\caption{Comparisons of our method with baselines on the NuScenes validation set. 
BEVFormer-t/s/b refers to BEVFormer-tiny/small/base in the BEVFormer's open-source codes. 'DFA3D' refers to the adaptation of 2D deformable attention into a (depth-weighted) 3D deformable attention within the BEVFormer model, as adopted in~\cite{dfa3d}.}
\label{Tab:comparison}
\vspace{-1em}
\end{table*}

\subsection{Experimental Setup}

\noindent \textbf{Dataset and Metrics} We evaluate our MvAcon on the challenging large-scale NuScenes dataset~\cite{caesar2020nuscenes} and Waymo dataset~\cite{waymo}. The \textbf{NuScenes} dataset includes 1,000 scene sequences, which are divided into training, validation, and testing subsets in a 700/150/150 split. Each sequence in the NuScenes dataset is a 20-second video clip, annotated at a rate of 2 frames per second (FPS). The NuScenes dataset employs a comprehensive suite of evaluation metrics for assessing detection performance. These metrics comprise mean Average Precision (mAP), mean Average Translation Error (mATE), mean Average Scale Error (mASE), mean Average Orientation Error (mAOE), mean Average Velocity Error (mAVE), mean Average Attribution Error (mAAE), and the NuScenes Detection Score (NDS). The \textbf{Waymo} dataset contains 798 training sequences and 202 validation sequences. We use a subset of the training set (Waymo-mini) by sampling every third frame from the training sequences following \cite{bevformer}.

\noindent \textbf{Implementation Details} We leverage open-source code bases (PETR~\cite{petr}, BEVFormer~\cite{bevformer}, and DFA3D~\cite{dfa3d}) in our experiments. To ensure a fair and stringent comparison, we maintain all original configurations of these methods, making only one modification: the addition of an attentive contextualization module. We conduct qualitative analysis and ablation study on the BEVFormer-base model. We train all models for 24 epochs using 8 NVIDIA Tesla A100 GPUs, following the configurations and settings outlined in previous works~\cite{bevformer, dfa3d, petr}.

\subsection{The Effectiveness of MvACon across Different Methods} 

To demonstrate the effectiveness of our proposed MvACon method, we first apply it to two typical query-based MV3D object detection paradigms: the perspective-based decoder-only detector (PETR~\cite{petr}) and the encoder-decoder based detector (BEVFormer~\cite{bevformer}). We choose these as our baselines because state-of-the-art (SOTA) query-based MV3D detectors~\cite{dfa3d, bevformerv2, liu2023petrv2, streampetr} primarily follow these two paradigms. We also test our method on DFA3D~\cite{dfa3d} to demonstrate its generalizability to SOTA methods.

On the NuScenes dataset, Table~\ref{Tab:comparison} shows that our proposed MvACon consistently improves performance across different detectors. Specifically, for the perspective-based decoder-only detector PETR, it improves the baseline by 0.8 NDS. For the encoder-decoder based detector BEVFormer, our method achieves an improvement of 1.3 in NDS on average. On a more advanced, depth-context enhanced BEVFormer (DFA3D), our method further improves performance by up to 0.5 NDS. Notably, our MvACon achieves significant improvement in location (mAP, mATE), orientation (mAOE), and velocity prediction (mAVE) for encoder-decoder based detectors. It also markedly enhances performance in location (mAP, mATE) and velocity (mAVE) prediction for the perspective-based decoder-only detector. 

\begin{table}[h]
% \vspace{-1em}
\begin{center}
\resizebox{0.9\linewidth}{!}{
\begin{tabular}{l|cc}
\toprule
\textbf{Method} & \textbf{LET-mAPL}$\uparrow$ & \textbf{LET-mAPH}$\uparrow$ \\

\midrule
 
BEVFormer-ResNet101 \cite{bevformer} & 34.9 & 46.3 \\

\midrule

BEVFormer-ResNet101-MvACon & \textbf{35.7 (+0.8)} & \textbf{47.5 (+1.2)} \\

\bottomrule
\end{tabular}
}
\end{center}
\vspace{-1.5em}
\caption{Comparisons on the Waymo-mini.}
\label{Tab:waymo}\vspace{-4mm}
\end{table}

On the Waymo dataset, since there are few released codes for MV3D detectors on Waymo except for the BEVFormer, we only test BEVFormer on Waymo-mini following its settings with results shown in Table~\ref{Tab:waymo}. Our MvACon shows consistent improvement on Waymo metrics.

\begin{figure*}[ht]
    \centering
    \includegraphics[width=0.85\textwidth]{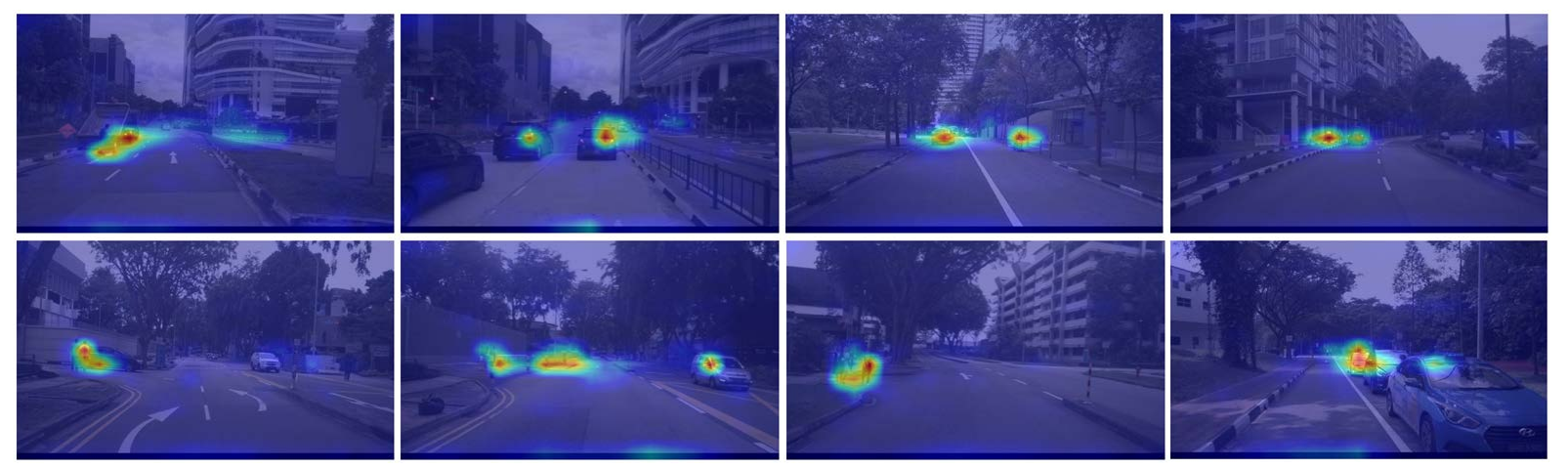}
    \vspace{-1em}
    \caption{Visualization results of learned cluster contexts in our MvACon on the NuScenes validation set. We sum all the learned clusters along the channel and upsample it to the original image resolution through bilinear interpolation. We observed that the learned cluster context encodes abundant context information in the scene. We provide details with raw images in the supplementary.}
    \label{fig:hm}
    \vspace{-1em}
\end{figure*}

\begin{figure*}[ht]
    \centering
    \includegraphics[width=0.88\textwidth]{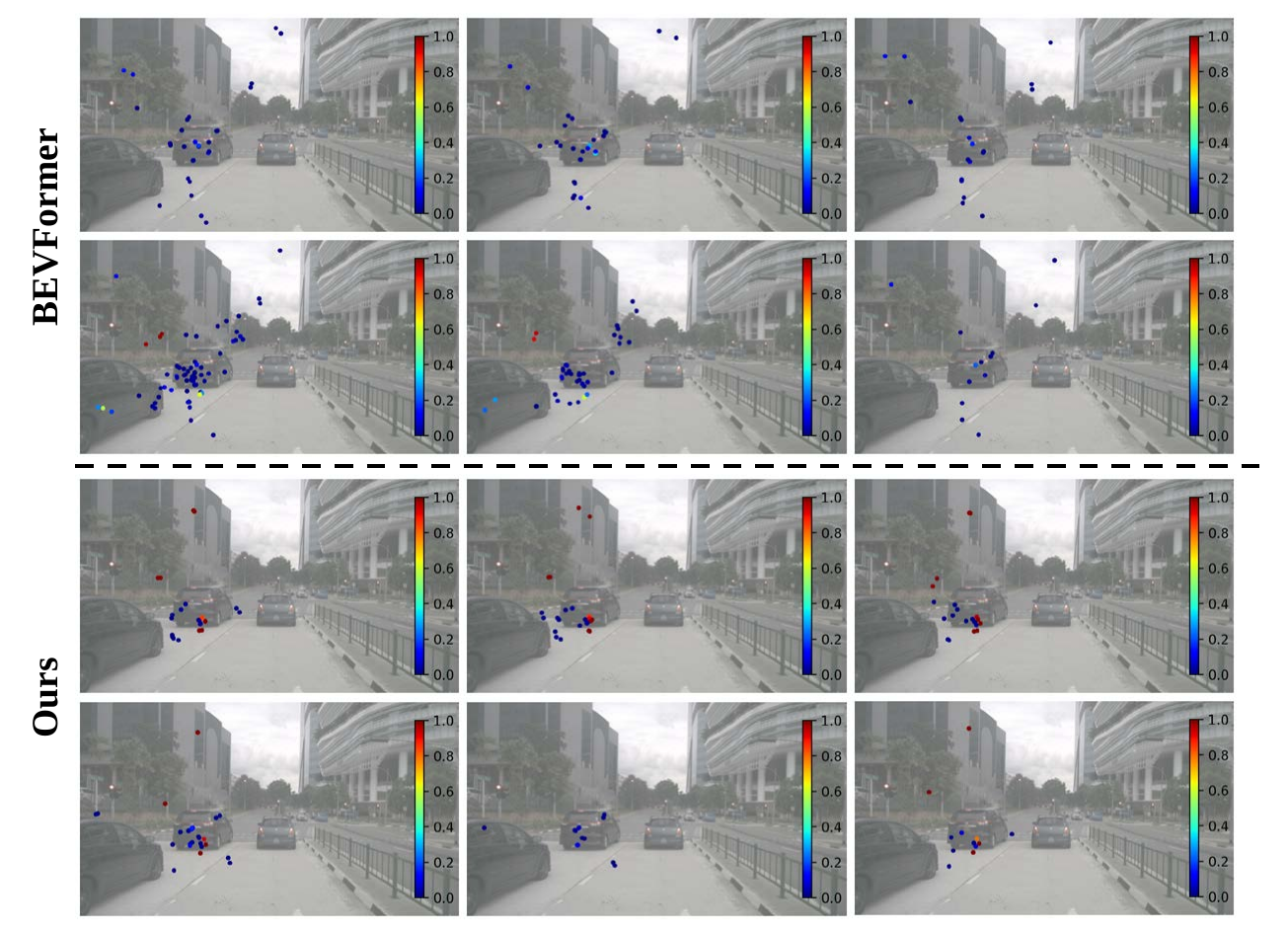}
    \vspace{-1em}
    \caption{Visualization results of the deformable points originating from a 2D reference point, which is projected from a 3D BEV anchor point in the BEVFormer encoder, on NuScenes validation set. We utilize the same BEV anchor point as demonstrated in Fig.~\ref{fig:teaser}. From left to right and up to bottom, we display the deformable points output from each layer (\#1-\#6) in the encoder, respectively.}
    \label{fig:dfpts}
    \vspace{-1em}
\end{figure*}

\begin{figure*}[ht]
    \centering
    \includegraphics[width=0.95\textwidth]{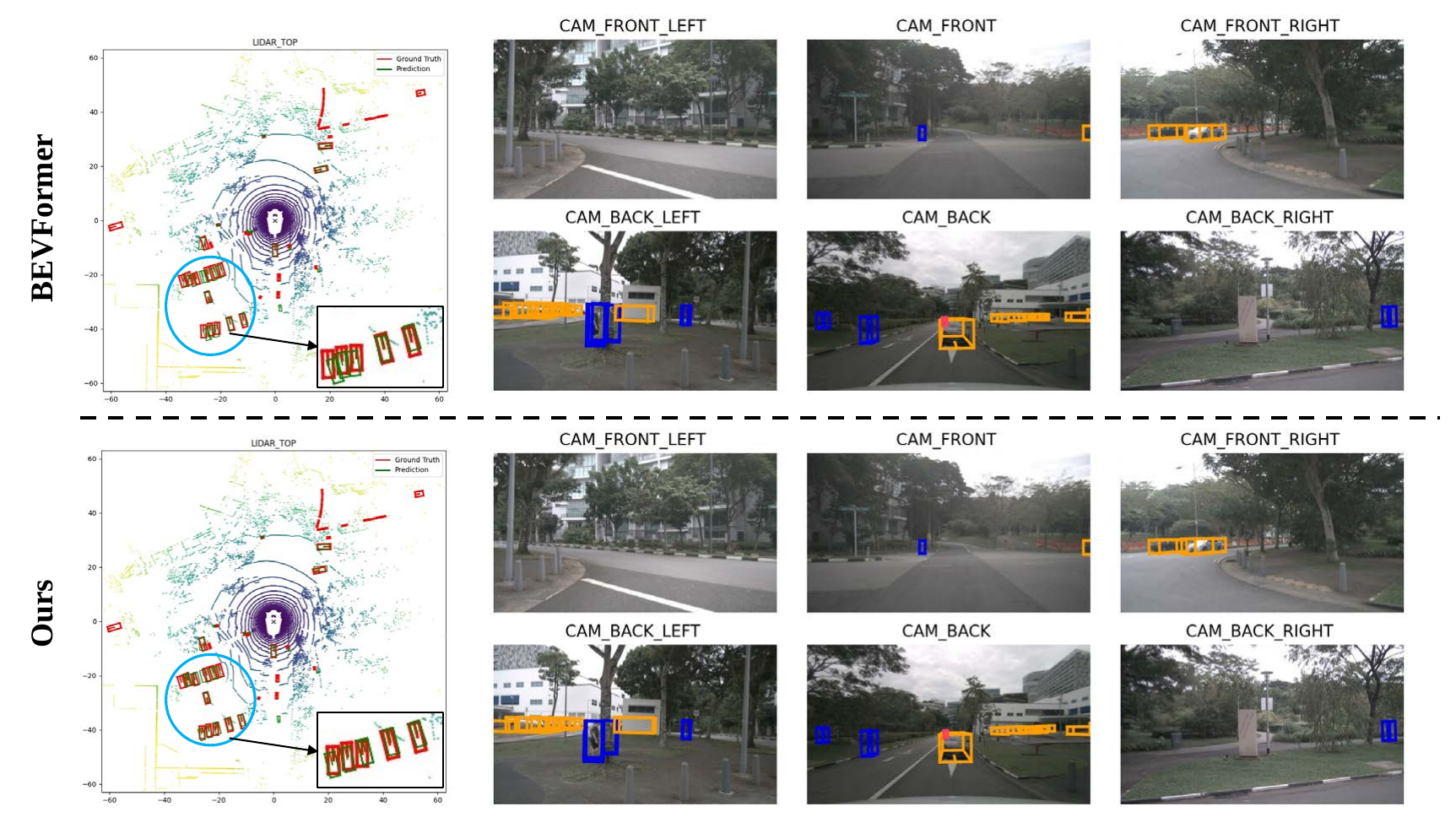}
    \vspace{-1em}
    \caption{Qualitative comparisons between BEVFormer and our MvACon method on NuScenes validation set.}
    \label{fig:qual2}
    \vspace{-1em}
\end{figure*}

\subsection{How Does Our MvACon Work?}

We elaborate on the effects of our MvACon by providing detailed analyses during the 2D-to-3D feature lifting process. We first demonstrate \textbf{what the learned cluster contexts encode}, then show \textbf{how these contexts affect the behavior of deformable points during feature lifting}. Lastly, we illustrate \textbf{how our MvACon improves detection results} by presenting a qualitative comparison within a scene. We select BEVFormer-base as our analysis target due to its incorporation of six layers of deformable attention modules in the encoder. More qualitative analysis is provided in the supplementary materials.

\vspace{0.3em }
\noindent \textbf{What do the Learned Cluster Contexts Encode?} We visualize the learned cluster context in a heatmap, by summing all clusters along the channel and then upsampling to the original image resolution using bilinear interpolation. The results, shown in Fig.~\ref{fig:hm}, reveal that despite the low resolution of the original summed heatmap, it can accurately locate foreground objects in the scene after upsampling. This suggests that the cluster contexts encode the foreground layouts in a scene. Furthermore, we note that the upsampled heatmap shows a high response to various foreground objects, even when they are spatially close in 2D. This is attributed to the effectiveness of our MvACon in dense scenes, as demonstrated and analyzed later in our experiments.

\vspace{0.3em }
\noindent \textbf{How do Cluster Contexts Affect the Behavior of Deformable Points?} We demonstrate the dynamics of deformable points in Fig.~\ref{fig:dfpts}. These points originates from one 2D reference point, which is projected from a 3D BEV anchor point in the BEVFormer encoder. As the encoding layers deepen in the vanilla BEVFormer, we observe that most deformable points have low attention weights, suggesting the model's uncertainty about the relevance of the selected context. Consequently, these contexts contribute minimally during the encoding of BEV query features in the feature lifting process. In contrast, deformable points predicted by our method maintain high confidence weights on foreground objects (e.g., cars), as well as on surrounding buildings. We also note that points on buildings remain stable across encoding layers and consecutive frames (see the supplementary). These high-confidence deformed points in our MvACon may foster a local object-context aware coordinate system, enhancing overall performance, including the estimation of velocity and orientation. This observation aligns with our quantitative findings in Table \ref{Tab:comparison}.

\vspace{0.3em}
\noindent \textbf{Qualitative Comparisons with the Baseline Method.} We provide qualitative comparison with BEVFormer in Fig.~\ref{fig:qual2}. Our method exhibits superior performance in dense scenarios where objects are crowded and challenging to localize. We attribute this enhanced performance to the rich foreground layout context in the scene, facilitated by our attentive contextualization module. Additional qualitative comparisons are available in the supplementary.

\subsection{Ablation Studies}

\noindent \textbf{Effectiveness of Different Contextualization Methods.} Table \ref{Tab:contextualization} demonstrates the impact of various contextual methods on detection performance. We selected three representative contexts: local shift window-based context (SWin~\cite{liu2021swin}), global pooling-based context (PVTV2-linear~\cite{wang2022pvt}), and global cluster-based context (PaCa~\cite{grainger2023paca}). The results show that the local shift window-based context offers minimal improvements in detection performance, which can be attributed to similar local contexts already provided by convolutional backbone networks. Conversely, enhancing the original feature with global contexts, as observed in PVTV2-linear and PaCa experiments, leads to better performance. Notably, orientation and velocity show significant improvements with these global contexts. The inclusion of cluster contexts further enhances improvements in location, orientation, and velocity prediction, as evidenced in the global cluster experiment.

\begin{table}[h]
\vspace{-0.5em}
\begin{center}
\resizebox{0.98\linewidth}{!}{
\begin{tabular}{l|c|ccccc|c}
\toprule
\textbf{Method} & \textbf{mAP}$\uparrow$ & \textbf{mATE}$\downarrow$ & \textbf{mASE}$\downarrow$ & \textbf{mAOE}$\downarrow$ & \textbf{mAVE}$\downarrow$ & \textbf{mAAE}$\downarrow$ & \textbf{NDS}$\uparrow$ \\

\midrule
 
BEVFormer-b~\cite{bevformer} & 41.6 & 67.3 & \textbf{27.4} & 37.2 & 39.4 & 19.8 & 51.7 \\

\midrule

Shift Window (SWin~\cite{liu2021swin}) & 41.5 (-0.1) & 67.1 & 27.8 & 37.8 & 39.6 & 20.1 & 51.5 (-0.2) \\

Global Pooling (PVTV2-linear~\cite{wang2022pvt}) & 41.6 (+0.0) & 66.5 & 27.5 & 36.8 & 37.6 & \textbf{19.6} & 52.0 (+0.3) \\

Global Cluster (PaCa~\cite{grainger2023paca}) & \textbf{42.6 (+1.0)} & \textbf{66.4} & 27.6 & \textbf{35.0} & \textbf{36.2} & 20.0 & \textbf{52.8 (+1.1)} \\

\bottomrule
\end{tabular}
}
\end{center}
\vspace{-1.5em}
\caption{Ablation study on different contextualization methods.}

\label{Tab:contextualization} \vspace{-2mm}
\end{table}

\begin{table}[h]
\vspace{-0.5em}
\begin{center}
\resizebox{0.98\linewidth}{!}{
\begin{tabular}{l|c|ccccc|c}
\toprule
\textbf{Context} & \textbf{mAP}$\uparrow$ & \textbf{mATE}$\downarrow$ & \textbf{mASE}$\downarrow$ & \textbf{mAOE}$\downarrow$ & \textbf{mAVE}$\downarrow$ & \textbf{mAAE}$\downarrow$ & \textbf{NDS}$\uparrow$ \\

\midrule
 
Local (BEVFormer-b~\cite{bevformer}) & 41.6 & 67.3 & 27.4 & 37.2 & 39.4 & 19.8 & 51.7 \\

\midrule

Global Cluster & 41.6 & 66.4 & 27.4 & 38.3 & 35.1 & 19.5 & 52.1 \\

Local + Global Cluster & \textbf{42.6 (+1.0)} & \textbf{66.4} & 27.6 & \textbf{35.0} & \textbf{36.2} & 20.0 & \textbf{52.8 (+1.1)} \\

\bottomrule
\end{tabular}
}
\end{center}
\vspace{-1.5em}
\caption{Ablation study on the relationship between local contexts and global cluster-based contexts.}
\label{Tab:residual} \vspace{-2mm}
\end{table}

\begin{table}[h]
\vspace{-0.5em}
\begin{center}
\resizebox{0.98\linewidth}{!}{
\begin{tabular}{l|ccc|c|ccccc|c}
\toprule
\textbf{Method} & \#layers & \#clusters & Cross-level & \textbf{mAP}$\uparrow$ & \textbf{mATE}$\downarrow$ & \textbf{mASE}$\downarrow$ & \textbf{mAOE}$\downarrow$ & \textbf{mAVE}$\downarrow$ & \textbf{mAAE}$\downarrow$ & \textbf{NDS}$\uparrow$ \\

\midrule
 
BEVFormer-b~\cite{bevformer} & - & - & - & 41.6 & 67.3 & 27.4 & 37.2 & 39.4 & 19.8 & 51.7 \\

\midrule

Baseline 1 & 3 & 100 & \checkmark & 41.1 & \textbf{65.6} & \textbf{27.1} & 37.7 & 36.4 & 18.9 & 52.0 \\

Baseline 2 & 6 & 50 & \checkmark & 41.3 & 66.3 & 27.3 & 36.1 & 34.1 & \textbf{18.8} & 52.4 \\

Baseline 3 & 6 & 100 & - & 42.4 & 66.8 & 27.6 & 37.1 & 36.3 & {19.1} & 52.5 \\

MvACon & 6 & 100 & \checkmark & \textbf{42.6} & {66.4} & 27.6 & \textbf{35.0} & \textbf{36.2} & 20.0 & \textbf{52.8} \\

\bottomrule
\end{tabular}
}
\end{center}
\vspace{-1.5em}
\caption{Ablation study on the structure of our attentive contextualization module.}

\label{Tab:cluster}\vspace{-2mm}
\end{table}
\vspace{-0.5em}
\begin{table}[h]
\begin{center}
\resizebox{0.98\linewidth}{!}{
\begin{tabular}{c|c|ccccc|c}
\toprule
\textbf{Clustering Method} & \textbf{mAP}$\uparrow$ & \textbf{mATE}$\downarrow$ & \textbf{mASE}$\downarrow$ & \textbf{mAOE}$\downarrow$ & \textbf{mAVE}$\downarrow$ & \textbf{mAAE}$\downarrow$ & \textbf{NDS}$\uparrow$ \\

\midrule
 
BEVFormer-b~\cite{bevformer} & 41.6 & 67.3 & 27.4 & 37.2 & 39.4 & 19.8 & 51.7 \\

\midrule

Linear & 41.9 & \textbf{65.6} & \textbf{27.2} & 38.2 & \textbf{34.8} & 19.3 & 52.4 \\

MLP & 42.4 & 66.4 & 27.3 & 38.2 & 35.0 & \textbf{19.2} & 52.6 \\

Conv & \textbf{42.6} & 66.4 & 27.6 & \textbf{35.0} & {36.2} & 20.0 & \textbf{52.8} \\

\bottomrule
\end{tabular}
}
\end{center}
\vspace{-1.5em}
\caption{Ablation study on clustering operations in the attentive contextualization module.}

\label{Tab:cluster_methods}\vspace{-2mm}
\end{table}

\vspace{0.3em}
\noindent \textbf{Relationship between Local Contexts and Global Cluster Contexts.} Table \ref{Tab:residual} reveals the relationship between local and global cluster contexts in enhancing feature learning for view lifting. The exclusive use of global cluster context results in improved velocity prediction, while local attention contributes to better orientation prediction results. Combining these two contexts enhances predictions in location, orientation, and velocity. This highlights the complementary role that global cluster contexts play in feature encoding for view lifting.

\noindent \textbf{Structure of the Attentive Contextualization Module.} Table~\ref{Tab:cluster} shows the impact of the structure of our attentive contextualization method. Baseline 1 demonstrates that attentive contextualization can efficiently encode feature information. With only three layers, MvACon achieves an improvement of 0.3 NDS compared to the vanilla BEVFormer. Baseline 2 indicates that attentive contextualization requires a sufficient number of clusters to extract abundant cluster contexts in the scene. Baseline 3 suggests that attending to clusters across the feature map aids in improving orientation, velocity, and mAP prediction. Table~\ref{Tab:cluster_methods} illustrates that the convolution operation for clustering yields better results in terms of orientation, mAP, and NDS prediction. Conversely, point-based operations (such as a linear layer or multi-layer perceptron) demonstrate superior performance in location and velocity prediction.

\section{Conclusion}

This paper presents Multi-View Attentive Contextualization (MvACon) for improving  query-based multi-view 3D (MV3D) object detection. It addresses the limitations of two main paradigms of query based MV3D object detector in a unified way: decoder-only dense attention methods like PETR, which lack high-resolution features due to computational constraints, and encoder-decoder sparse 2D/3D deformable attention methods such as BEVFormer and DFA3D. Our MvACon contextualizes the original feature maps extracted from the backbone network using a cluster-attention operation built on the recently proposed patch-to-cluster attention (PaCa).
In experiments, we show that our MvACon effectively and consistently enhances query-based MV3D object detectors by encoding more useful contexts, thereby facilitating better 2D-to-3D feature lifting. Rigorously controlled experiments reveal that, for decoder-only detectors, the cluster attention contextualization significantly improves localization and velocity prediction. For encoder-decoder based detectors, it effectively reduces errors in location, orientation, and velocity.
\section*{Acknowledgments}
X. Liu and T. Wu were partly supported by NSF IIS-1909644,  
 ARO Grant W911NF1810295, ARO Grant W911NF2210010, NSF CMMI-2024688, NSF IUSE-2013451, and a research gift fund from the Innopeak Technology, Inc. (an affiliate of OPPO).  
The views and conclusions contained herein are those of the authors and should not be interpreted as necessarily representing the official policies or endorsements, either expressed or implied, of the ARO, NSF, or the U.S. Government. The U.S. Government is authorized to reproduce and distribute reprints for Governmental purposes not withstanding any copyright annotation thereon.

\clearpage

{
    \small
    \bibliographystyle{ieeenat_fullname}
    \bibliography{main}
}

\clearpage
\setcounter{page}{1}
\maketitlesupplementary
\setcounter{section}{0}
\setcounter{figure}{5}
\setcounter{table}{6}

\section*{Overview}
In this supplementary material, we provide more details on the following aspects that are not presented in the main paper due to space limit: 
\begin{itemize}
    \item \textit{Computation and memory cost} are provided in Sec.~\ref{sec:computation}.  
    \item \textit{Supplementary qualitative results on NuScenes validation split} are provided in Sec.~\ref{sec:more_qual}.  
\end{itemize}

\section{Computation and Memory Cost} \label{sec:computation}

\begin{table}[h]
\begin{center}
\vspace{-0.5em}
\resizebox{0.9\linewidth}{!}{
\begin{tabular}{l|ccc|c}
\toprule
\textbf{Method} & \textbf{Speed (FPS)} & \textbf{GPU Mem (MB)} & \textbf{\#Param (M)} & \textbf{NDS}$\uparrow$ \\

\midrule

PETR-VovNet-99~\cite{petr} & 9.8 & 3638 & 83.07 & 42.6 \\

PETR-VovNet-99-MvACon & 9.6 & 3638 & 84.75 & \textbf{43.4 (+0.8)} \\

\midrule

BEVFormer-b~\cite{bevformer} & 3.9 & 6928 & 69.14 & 51.7 \\

BEVFormer-b-MvACon-lite &  3.2 & 6936 & 70.75 & \textbf{52.5 (+0.8)} \\

BEVFormer-b-MvACon & 3.0 & 11452 & 70.75 & \textbf{52.8 (+1.1)} \\

\bottomrule
\end{tabular}
}
\end{center}
\vspace{-1.5em}
\caption{Efficiency and resource consumption of MvACon on PETR and BEVFormer. MvACon-lite refers to the model without using the concatenation of cluster contexts from all feature pyramids. This will greatly reduce extra GPU memory consumption, with only 0.3 NDS droppped compared with our full model.}
\label{Tab:cost}
\vspace{-1em}
\end{table}

Computation and memory cost of our MvACon is provided in Tab.~\ref{Tab:cost}. We use the parameter calculation script provided by BEVFormer's open source codebase: \url{https://github.com/fundamentalvision/BEVFormer}. 

Our MvACon is able to improve PETR with negligible computation cost. We tested two versions of BEVFormer-b-MvACon: a lite version and a full model. In the lite version, we enforce cluster attention within each feature pyramid level instead of using clusters from all levels. This will largely reduce the computation cost. It shows that our lite version is able to improve the baseline with only 8 MB extra GPU memory cost with 0.8 NDS improvement. Using our full model, we are able to improve the baseline with 1.1 NDS improvement. These results clearly demonstrates the effectiveness and necessity of incorporating useful contexts before feature lifting.

\section{More Qualitative Results on NuScenes} \label{sec:more_qual}

\begin{figure*}[ht]
    \centering
    \includegraphics[width=0.9\textwidth]{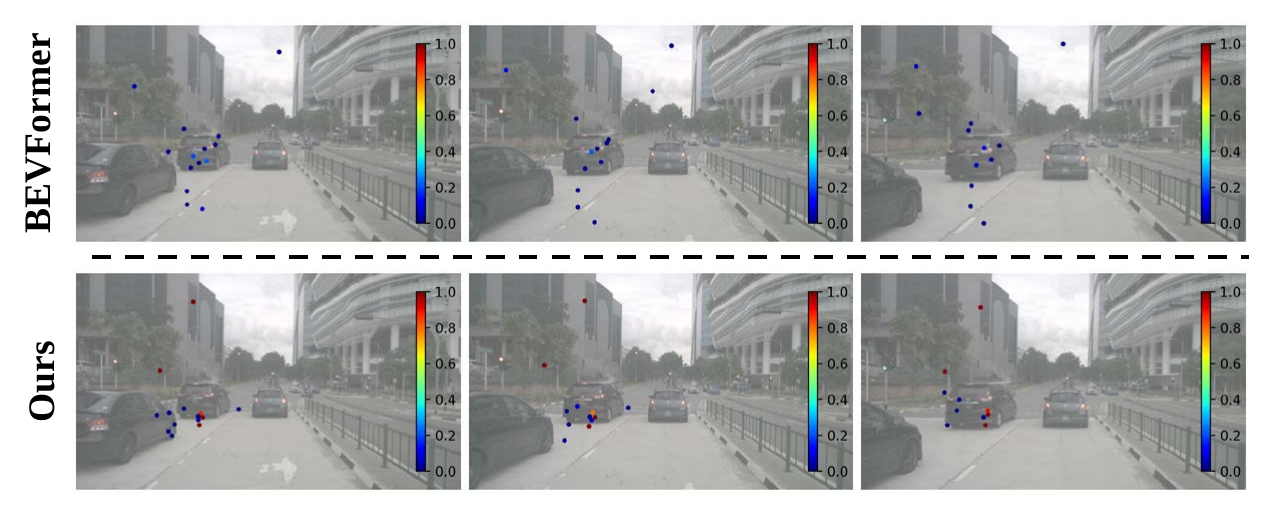}
    \vspace{-1em}
    \caption{Visualization results of the deformable points originating from a 2D reference point across 3 consecutive frames on NuScenes validation set. This 2D reference point is projected from a 3D BEV (Bird's Eye View) anchor point in the BEVFormer encoder. We use the same BEV anchor point as the one presented in our main paper. From left to right, we exhibit the deformable points outputted from the encoder's final layer, arranged in chronological order ($t$-1, $t$, $t$+1).}
    \label{fig:qual_time}
    % \vspace{-2em}
\end{figure*}

\noindent \textbf{Qualitative results for deformable points across consecutive frames.} We visualize the deformable points across 3 consecutive frames in Fig.~\ref{fig:qual_time}. We observe that our MvACon is able to learn stable and meaningful high-response deformable points on both cars and surrounding buildings.

\begin{figure*}[ht]
    \centering
    \includegraphics[width=0.8\textwidth]{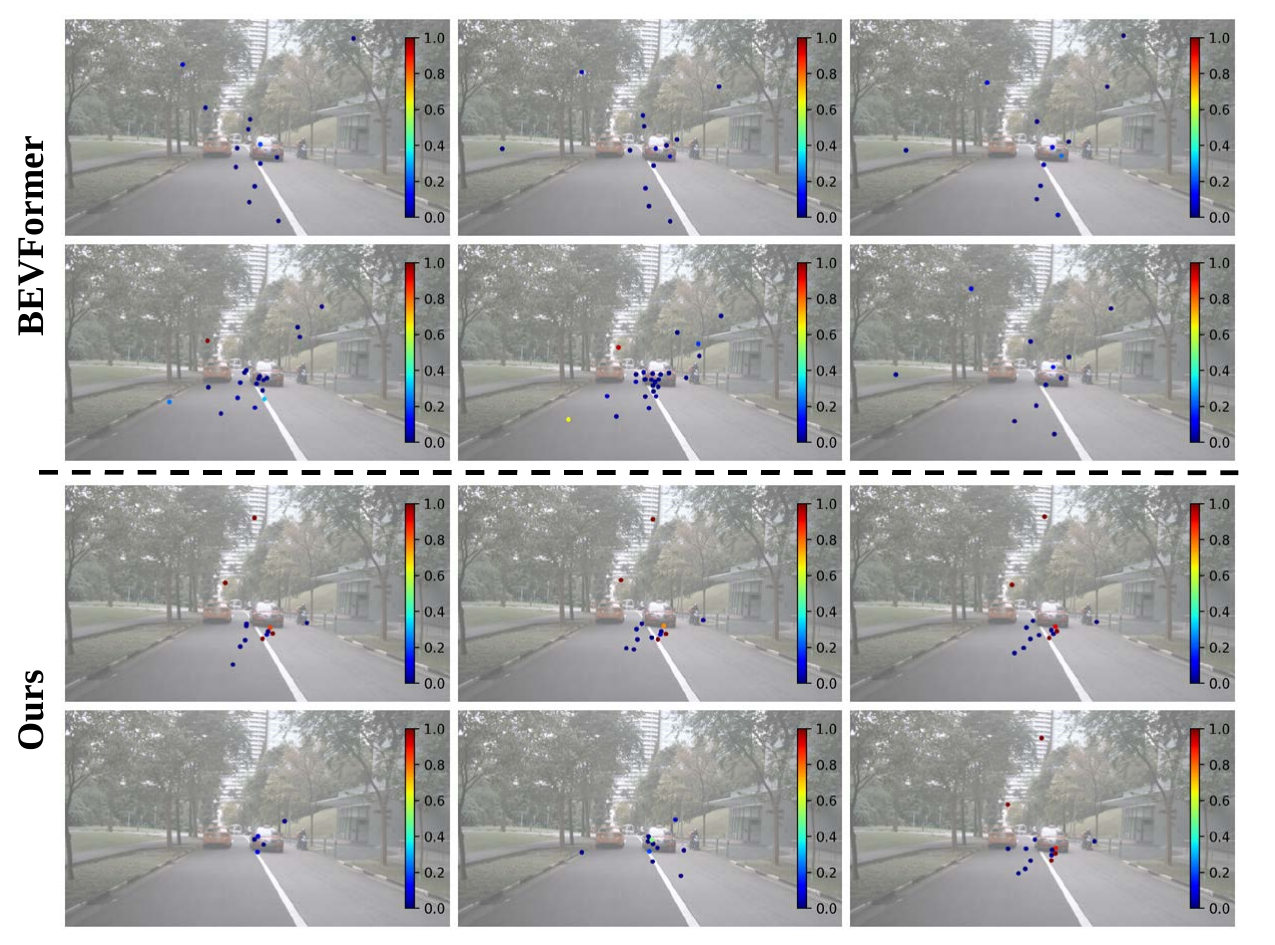}
    \vspace{-1em}
    \caption{Visualization results of the deformable points originating from a 2D reference point, which is projected from a 3D BEV anchor point in the BEVFormer encoder, on NuScenes validation set. We utilize the a BEV anchor point one the right car. From left to right and up to bottom, we display the deformable points output from each layer (\#1-\#6) in the encoder, respectively.}
    \label{fig:supp_df1}
    % \vspace{-2em}
\end{figure*}

\begin{figure*}[ht]
    \centering
    \includegraphics[width=0.8\textwidth]{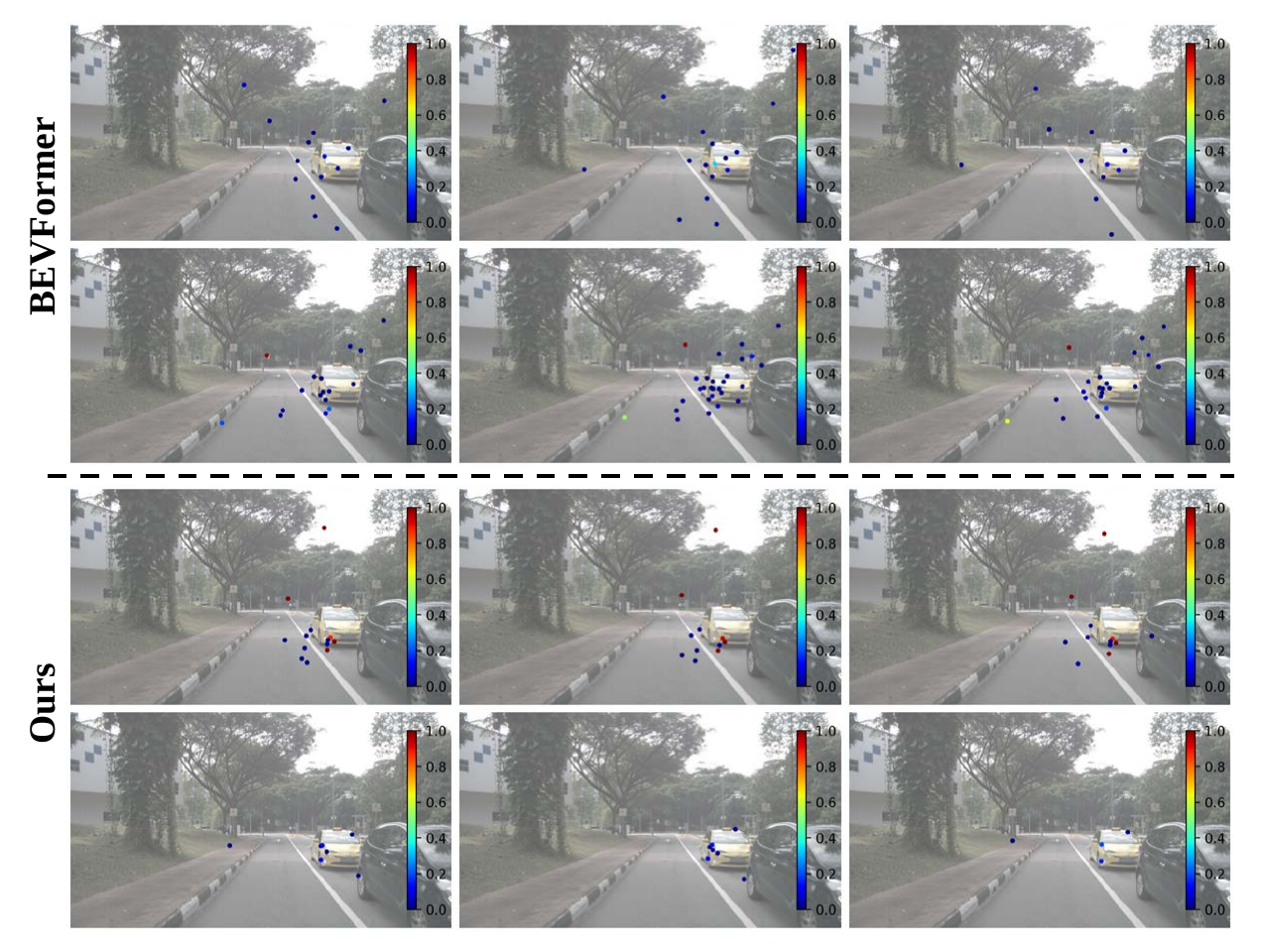}
    \vspace{-1em}
    \caption{Visualization results of the deformable points originating from a 2D reference point, which is projected from a 3D BEV anchor point in the BEVFormer encoder, on NuScenes validation set. We utilize the a BEV anchor point one the yellow car. From left to right and up to bottom, we display the deformable points output from each layer (\#1-\#6) in the encoder, respectively.}
    \label{fig:supp_df2}
    % \vspace{-2em}
\end{figure*}

\noindent \textbf{Supplementary qualitative results for deformable points in different scenes.} We visualize the deformable points in different scenes in Fig.~\ref{fig:supp_df1} and Fig.~\ref{fig:supp_df2}. We observe that our MvACon is able to learn meaningful high-response deformable points on cars and surrounding references, which could be helpful in improving the prediction of object location, orientation and velocity.  

\begin{figure*}[ht]
    \centering
    \includegraphics[width=0.98\textwidth]{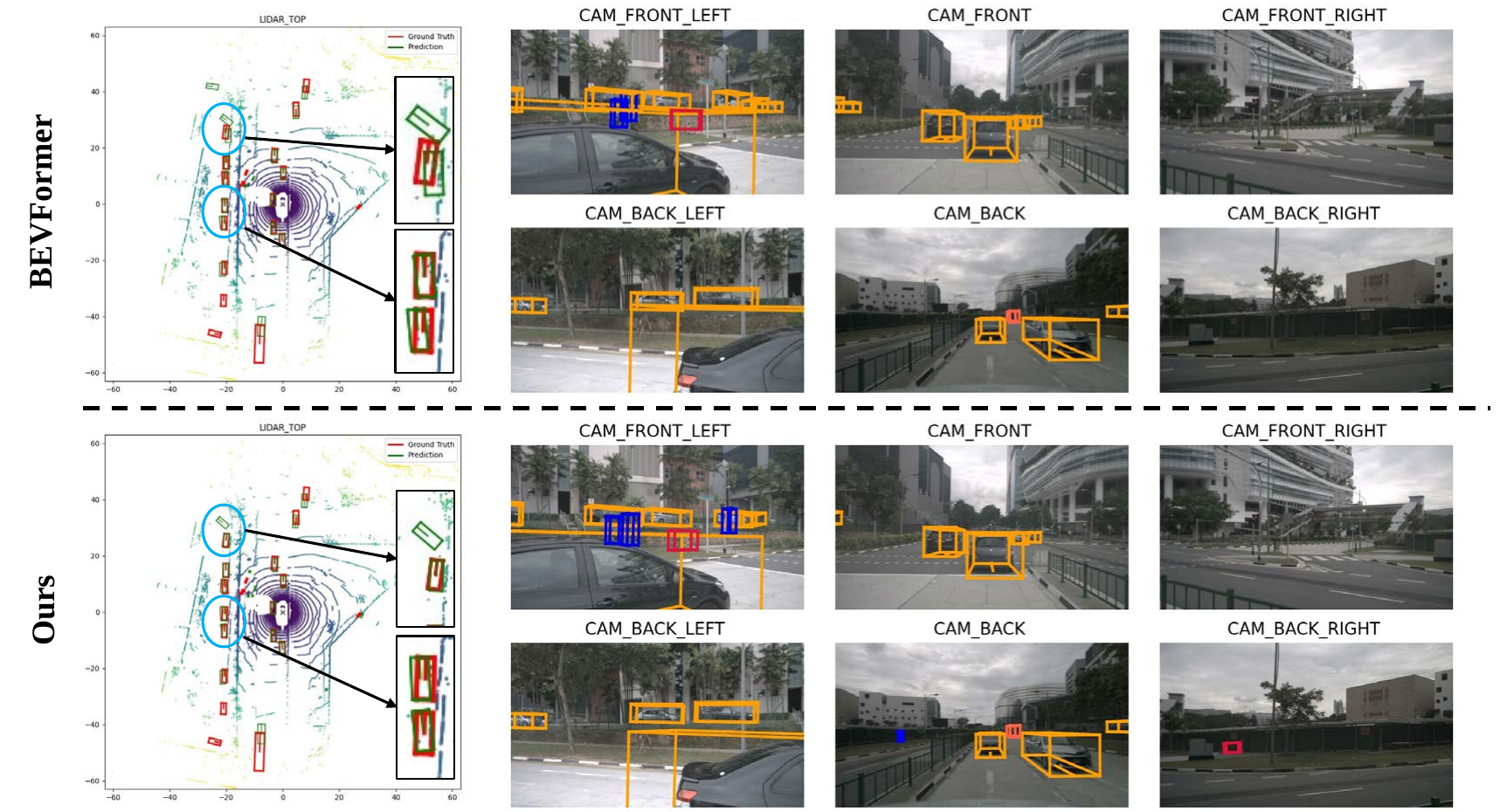}
    \vspace{-1em}
    \caption{Qualitative comparisons between BEVFormer and our MvACon method on NuScenes validation set.}
    \label{fig:supp_qual1}
    % \vspace{-2em}
\end{figure*}

\begin{figure*}[ht]
    \centering
    \includegraphics[width=0.98\textwidth]{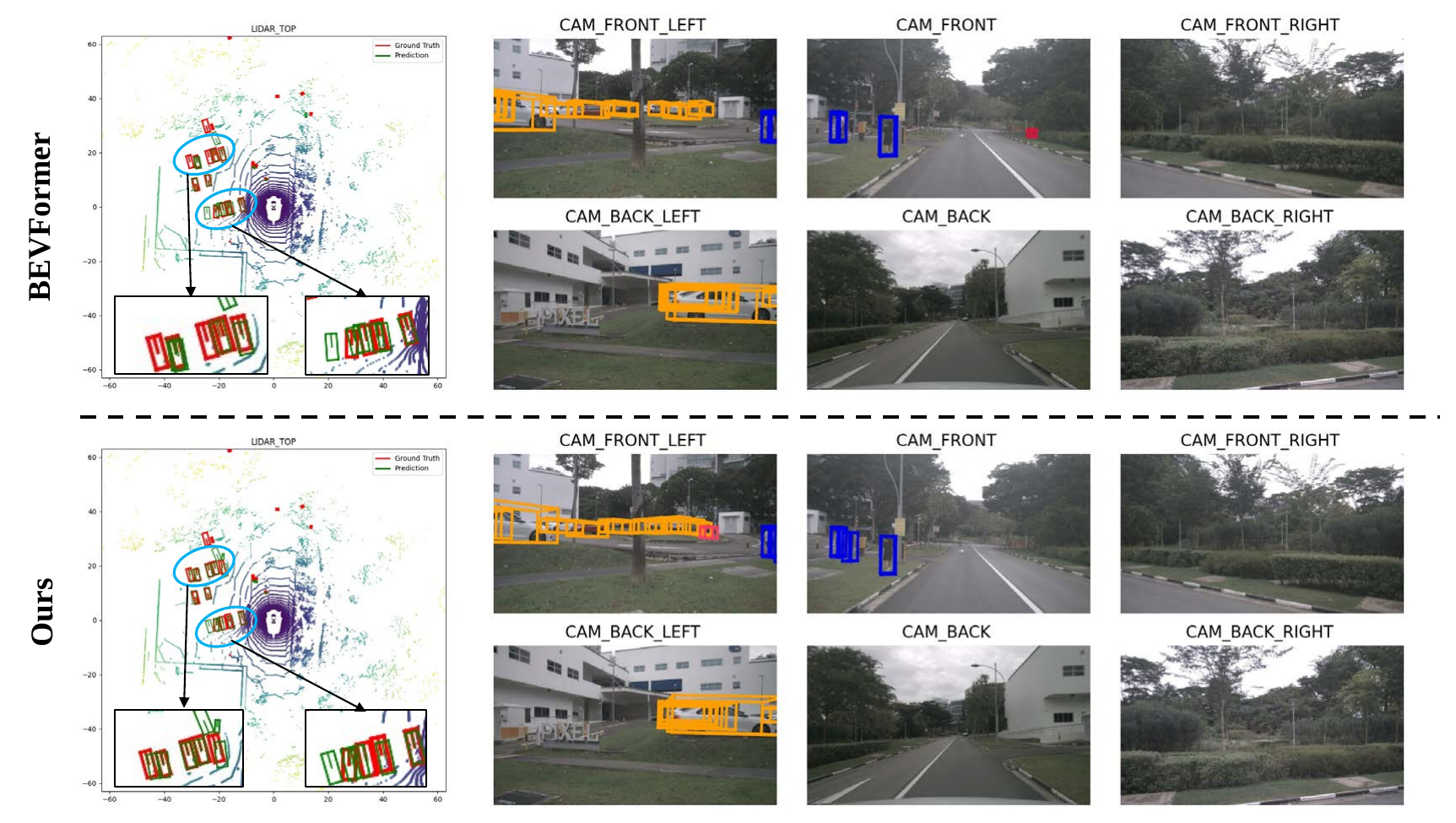}
    \vspace{-1em}
    \caption{Qualitative comparisons between BEVFormer and our MvACon method on NuScenes validation set.}
    \label{fig:supp_qual3}
    % \vspace{-2em}
\end{figure*}

\begin{figure*}[ht]
    \centering
    \includegraphics[width=0.98\textwidth]{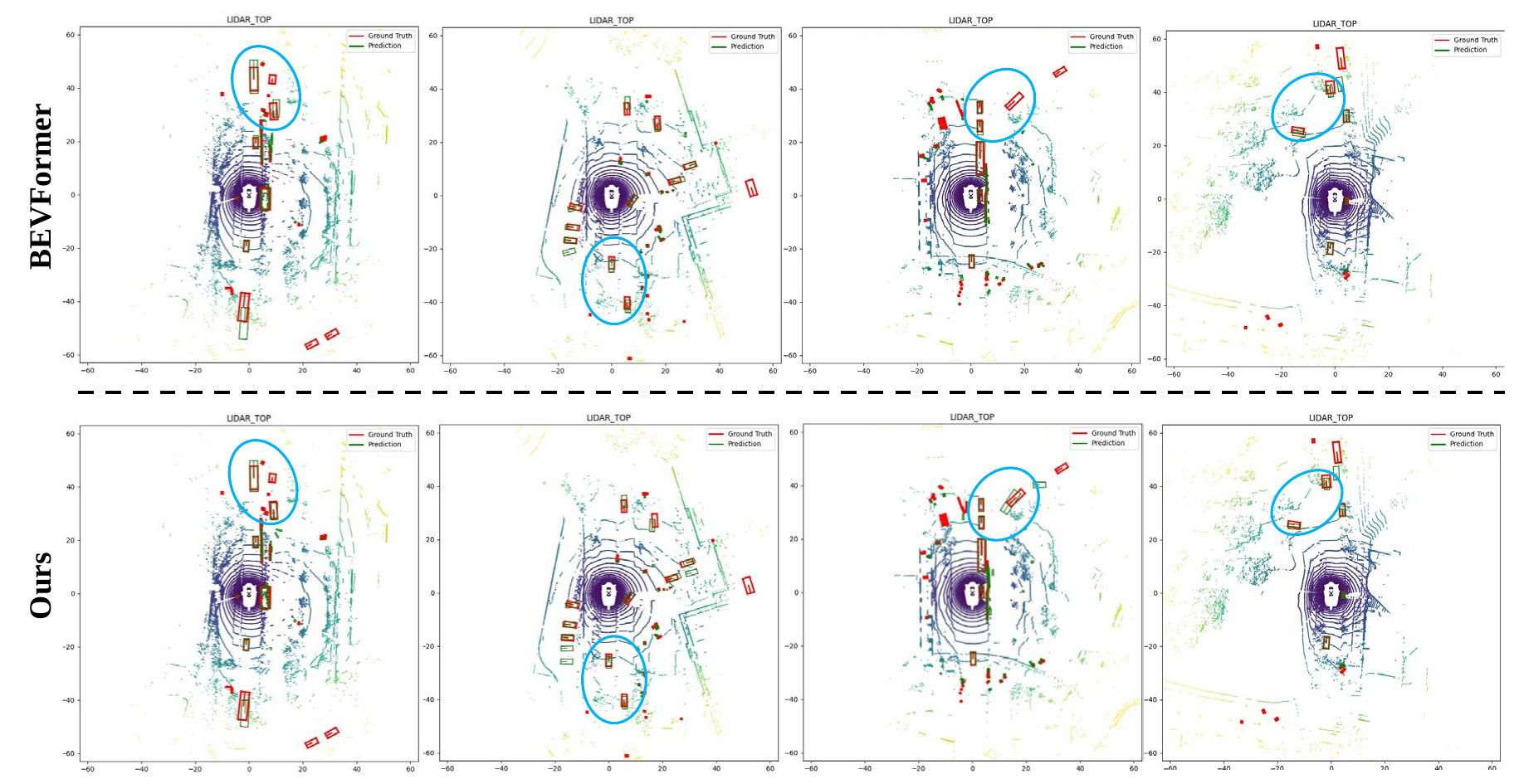}
    \vspace{-1em}
    \caption{Qualitative comparisons between BEVFormer and our MvACon method on NuScenes validation set.}
    \label{fig:supp_qual4}
    % \vspace{-2em}
\end{figure*}

\noindent \textbf{Supplementary qualitative comparison for detection results on NuScenes validation set.} We visualize prediction results on NuScenes validation set and compare it with BEVFormer in Fig.~\ref{fig:supp_qual1}, Fig.~\ref{fig:supp_qual3} and Fig.~\ref{fig:supp_qual4}. We observe that our MvACon performs better in dense scenes.

\begin{figure*}[ht]
    \centering
    \includegraphics[width=0.98\textwidth]{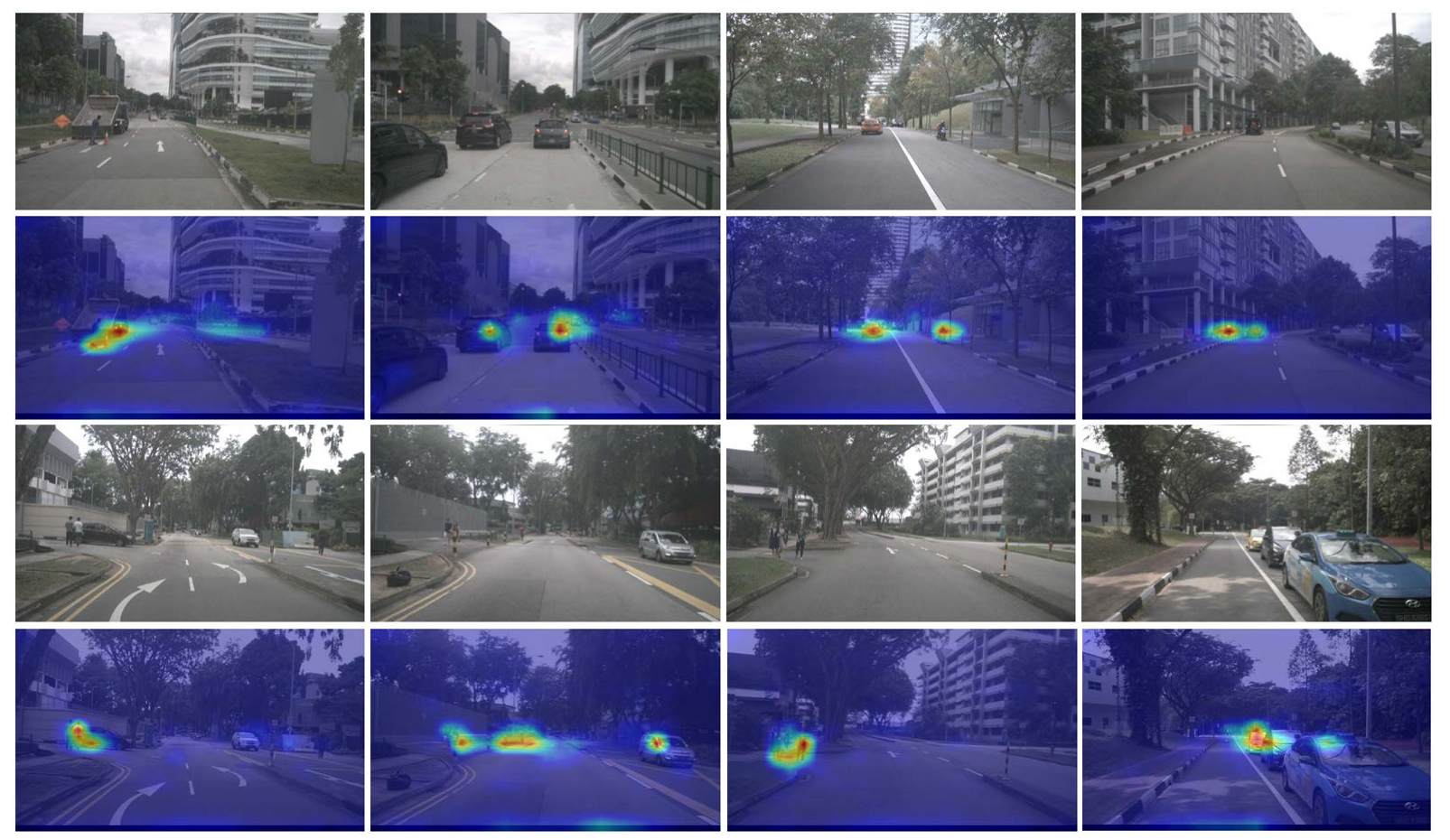}
    \vspace{-1em}
    \caption{Visualization results of learned cluster contexts with raw images in our proposed attentive contextualization module on NuScenes validation set. We sum all the learned clusters along the channel and upsample it to the original image resolution through bilinear interpolation. We observed that the learned cluster context encodes abundant context information in the scene.}
    \label{fig:supp_qual6}
    % \vspace{-2em}
\end{figure*}

\noindent \textbf{Supplementary visualization for learned cluster heatmap} We provide the detailed visualization results of learned cluster contexts with raw images in Fig.~\ref{fig:supp_qual6}. This uses the same scene shown in Fig. 3 of our main paper. The only difference is that we include raw images in supplementary materials. We observe that the learned cluster heatmap has high response on foreground contexts.

\end{document}